\documentclass[10pt]{article} 

\usepackage[accepted]{rlj} 

\usepackage{amssymb}            
\usepackage{mathtools}          
\usepackage{mathrsfs}           
\usepackage{graphicx}           
\usepackage{subcaption}         
\usepackage[space]{grffile}     
\usepackage{url}                
\usepackage{lipsum,amsmath,amsthm}             
\usepackage{adjustbox}
\theoremstyle{plain}

\theoremstyle{definition}

\theoremstyle{remark}

\usepackage{rlj}

\title{ASALT: Adaptive State Alignment for Lateral Transfer in Multi-agent  Reinforcement Learning}

\setrunningtitle{ASALT: Transfer Learning in MARL}

\author{Anurag Akula\textsuperscript{1}, Satheesh K. Perepu \textsuperscript{2}, Abhishek Sarkar \textsuperscript{2}, Kaushik Dey \textsuperscript{2}}

\emails{}

\affiliations{
$^{1}$\textbf{Indian Institute of Technology Madras, India}\\
$^{2}$\textbf{Ericsson Research, India}
}

\contribution{
A novel methodology for lateral information transfer between source and target domains in multi-agent reinforcement learning (MARL) is proposed.
}
{
Several prior studies on lateral transfer have proceeded under the assumption that there is no dimensionality mismatch between the observation and state spaces across the source and target domains.
}
\contribution{
Critic transfer provides measurable gains in learning efficiency and training stability across several evaluated settings. However, it does not universally outperform actor-only transfer, particularly in scenarios involving substantial increases in state-space dimensionality, where coordination strategies must be relearned
    
    }
    {
In the existing literature, information transfer has been considered exclusively between actors. We have demonstrated that incorporating information transfer between critics yields the most efficient configurations. 
    }


\summary{This paper introduces ASALT, a transfer-learning framework for multi-agent reinforcement learning that addresses a central limitation in existing techniques: the assumption that source and target domains share the same observation and state dimensionality. ASALT resolves this by using specialized adapters that translate target-domain observations and global states into a latent space compatible with pretrained source agents.
These aligned embeddings are fed through frozen source actors and critics, whose intermediate representations are laterally connected to the target actor and critic through a learnable weighting mechanism. This enables the target agents to leverage coordination strategies learned in the source domain without requiring structural similarity between domains.
The method supports upscaling, downscaling, and heterogeneous transfer. Extensive experiments across SMAC, MPE, and GRF environments show that ASALT not only accelerates learning but also mitigates negative transfer in settings where coordination requirements differ substantially. Several ablations further demonstrate the importance of joint adapter–target agent training, attention-based adapters, and critic-side transfer.
}

\begin{document}

\maketitle  

\begin{abstract}
Multi-agent reinforcement learning (MARL) addresses the problem of training multiple agents that pursue collaborative, competitive, or mixed objectives. Prior work has investigated transfer learning between source and target domains in MARL; however, the majority of existing approaches impose the constraint that the dimensionalities of the observation space and the global state space must be identical across domains.  In this paper, we introduce a method that explicitly accommodates mismatched state-space dimensionalities between source and target domains. The proposed approach, ASALT, incorporates both observation-level and state-level adapters that map the target-domain observations and global states into a shared embedding space, thereby enabling more effective transfer of knowledge across both actors and critics. These adapters can generate embeddings that support efficient strategy transfer across heterogeneous domains. Experimental results on multiple configurations in standard benchmark environments demonstrate that ASALT surpasses existing baselines in terms of sample efficiency and global return in cooperative settings, but its effectiveness depends on the degree of mismatch between source and target domains. Furthermore, our findings indicate that ASALT mitigates negative transfer, which frequently constitutes a major obstacle when transferring policies between domains with differing observation and action spaces. 
\end{abstract}

\section{Introduction}
\label{sec:intro} 

Multi-agent reinforcement learning (MARL) has been applied to diverse domains, including traffic management \citep{paper:MARL_App_Traffic}, power distribution \citep{paper:MARL_App_Power}, fleet management \citep{paper:MARL_App_Fleet}, autonomous driving \citep{paper:MARL_App_New}, and autonomous control \cite{paper:MARL_ATC}. Existing MARL algorithms learn either centralized policies \citep{paper:Deep_MAR_Cent2,paper:cent1} or decentralized policies \citep{paper:MARL_Dec,paper:MAPPO}. To improve decentralized policies, a widely used and effective paradigm is to employ centralized training during learning \citep{paper:IQL,paper:QMIX,paper:maven,paper:facmac}, known as \emph{Centralized Training and Decentralized Execution} (CTDE) \citep{paper:MARL_Review,paper:MARL_Review10,paper:deepRL_new}. In this work, we assume agents are trained using CTDE, which is well suited to realistic settings where agents must act autonomously at execution time without centralized information.

Conventional MARL transfer learning usually assumes a roughly static environment. In practice, however, related environments may differ in the number of agents or adversaries, even in competitive or mixed settings. In purely cooperative tasks, changes in the agent population modify the observation space and thus require significant adaptation in coordination transfer. Objectives may also shift from fully cooperative to mixed cooperative–competitive. In all these cases, humans seem to use prior experience to adapt quickly to new configurations, as when workers join or leave a factory team or the same staff is reassigned to more workstations than originally planned.

Methods for transferring knowledge from a source to a target domain in the context of single-agent RL have been investigated extensively. A subset of these approaches concentrates on learning mappings between observations and actions in the source and target domains, thereby circumventing the need to directly train a policy in the target domain \citep{paper:MARL_Review1,paper:DANN,paper:CORAL,paper:CycleGAN}. However, most such methods are designed for single-agent settings and are therefore not well suited for transferring coordination mechanisms, which are essential in multi-agent systems. 

Another line of work leverages concepts such as curriculum learning \citep{paper:EPC}, representation learning via embeddings \citep{paper:MARL_Transformer}, and lateral connections for transfer \citep{paper:MALT,paper:Ravisir}. Among these, lateral transfer–based methods are particularly notable, as they can accommodate a broad spectrum of transfer scenarios, including heterogeneous transfer as well as the scaling up or down of the number of agents.

Despite their advantages, lateral transfer–based techniques face a key limitation in real-world MARL: they usually assume identical observation-space dimensionality between source and target policies, an assumption rarely met in practice. This forces source and target policies to share the same architecture, greatly limiting applicability. In realistic settings, this becomes a major bottleneck. For instance, in the StarCraft II (SMAC) MARL benchmark \citep{paper:smac}, the observation-space dimensionality varies with map configuration and agent count, and similar issues occur in Google Research Football \citep{paper:google}. This mirrors human coordination, where changing the number of collaborators alters the field of observation and demands adapted coordination strategies.

In this paper, we propose an approach that can accommodate variations in the number of agents between source and target domains without imposing structural constraints on the network architecture. The proposed method employs adapter modules placed before the pre-trained source agents to transform the observations and global state (if available) in the target domain into latent embeddings. These embeddings encapsulate knowledge acquired from the source environments and are capable of conveying richer information, which, as evidenced by our experimental results, leads to more efficient training of the target agents.

The embeddings produced by the adapters are subsequently passed through the source agents, and the outputs from their intermediate layers are laterally transferred to the target agent via a weight-sharing mechanism. The target agent itself is trained using target-domain observations and states, enabling effective reuse of source representations while adapting to the new domain. In this work, we jointly train the adapters and the target agent to enhance the quality of the learned embeddings, which in turn supports effective transfer and helps mitigate negative transfer.

Empirical results on three benchmark environments, Starcraft II multi-agent challenge \citep{paper:smac}, Google Research Football \citep{paper:google} and Multi-particle environments \citep{paper:MPE}, demonstrate that the proposed ASALT method outperforms several baseline approaches with respect to both the training efficiency of target agents and the reduction of negative transfer.

\section{Related Works}
\label{sec:related}

Recently, there has been substantial interest in applying RL to a wide range of application domains. A key obstacle to the deployment of RL in real-world settings, however, is the challenge of generalization, namely the requirement that a learned agent remain effective under changes in the environment. Several classes of methods have been proposed to transfer knowledge from one domain to another, including (i) domain adaptation methods, (ii) transfer learning methods, and (iii) meta-learning methods.  

Domain adaptation methods focus on learning mappings between the state and action spaces of a source and a target domain, rather than directly learning a new policy in the target domain \citep{paper:DANN,paper:CORAL,paper:CycleGAN}. Transfer learning approaches primarily rely on fine-tuning a policy, initially trained in a source domain, to perform well in a target domain \citep{paper:RL_Transfer,paper:RL_Transfer1}. Meta-learning methods instead aim to train a shared model across multiple domains, often by explicitly incorporating domain information as an additional conditioning variable or parameter \citep{paper:metaRL}. All of these approaches predominantly address knowledge transfer in the context of single-agent RL.  

A direct extension of single-agent RL transfer techniques to MARL is non-trivial and typically does not yield efficient transfer. In MARL, one must not only transfer knowledge at the individual agent level but also preserve and adapt patterns of coordination among agents. Consequently, many single-agent transfer methodologies are not directly applicable. Within the MARL literature, several approaches have been proposed, including curriculum learning \citep{paper:EPC}, representation learning \citep{paper:MARL_Transformer}, and lateral transfer methods \citep{paper:MALT}. Some of these methods aim to transfer knowledge across domains by learning latent representations that serve as inputs to the policy. Curriculum learning methods seek to emulate human educational processes by training agents on tasks organized in order of increasing difficulty, for example by gradually increasing the number of agents. Another class of methods focuses on lateral transfer, where knowledge is propagated from one or more source domains to a target domain.  

Despite their advantages, these methods exhibit notable limitations. A prevalent assumption is that the dimensionality of the observation space remains identical across source and target domains. Furthermore, most existing approaches concentrate on transferring knowledge exclusively through the actors (policies), while largely overlooking the potential benefits of transferring information encoded in the critics (value functions). Since critics provide an efficient mechanism for estimating the advantage function, leveraging and transferring the information they contain could substantially improve learning efficiency in the target domain.

\section{Proposed ASALT approach}
\label{sec:proposed}

The proposed approach begins by considering the set of available trained source policies $\pi_{s^i}, \; i = 1,\dots,N$. The goal is to derive a joint policy in the target domain for a system comprising $M$ agents, leveraging knowledge transferred from the source domain. For clarity, each individual source policy operates over an observation space of dimension $\mathbf{o}_s$ and an action space of dimension $\mathbf{a}_s$, that is, $\pi_{s^i}:\mathbb{R}^{\mathbf{o}_s} \rightarrow \mathbb{R}^{\mathbf{a}_s}$. In addition, let $\mathbf{s}_s$ denote the dimensionality of the global state space in the source domain.  

We now consider the training of the target policies $\pi_{t^i}:\mathbb{R}^{\mathbf{o}_t} \rightarrow \mathbb{R}^{\mathbf{a}_t}$, where $\mathbf{o}_t$, $\mathbf{a}_t$ and $\mathbf{s}_t$ denote the dimensionalities of the observation, action and global state spaces of target domain respectively. In many environments, the dimensionalities of the observation space, global state space, and action space differ between the source and target domains, i.e., $\mathbf{o}_s \neq \mathbf{o}_t$, $\mathbf{s}_s \neq \mathbf{s}_t$, and $\mathbf{a}_s \neq \mathbf{a}_t$. Under such conditions, the source policies cannot be directly exploited for transfer learning because of the mismatches in the observation and global state spaces.  

To address this issue, we propose the introduction of adapter modules that explicitly compensate for the discrepancies in observation and global state representations, thereby enabling effective transfer of the source information to the target domain.

\begin{figure}[t]
    \centering
    \includegraphics[width=0.9\textwidth]{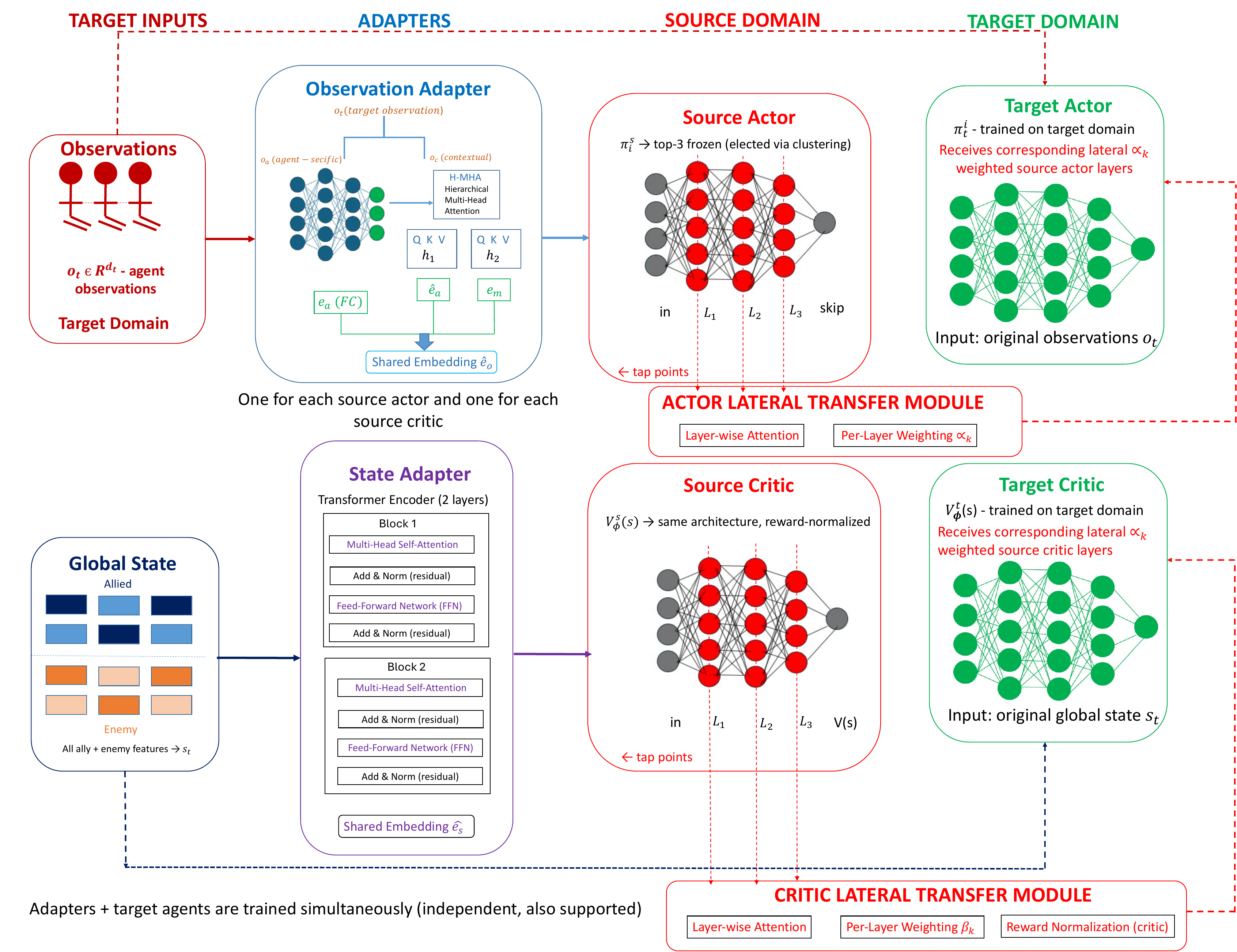}
    \caption{Overview of the proposed ASALT framework. 
(\textit{Top row}) Target-domain observations 
$o_t \in \mathbb{R}^{d_t}$ are factorized into 
agent-specific features $o_a$ and contextual features 
$o_c$, which are processed by a Hierarchical Multi-Head 
Attention (H-MHA)–based Observation Adapter to yield 
a shared embedding $\hat{e}_o$. This embedding is 
aligned with the input space of the frozen source 
actors $\pi^s_i$ (top-3, selected via clustering). 
(\textit{Bottom row}) The global state $s_t$, 
comprising both allied and enemy feature representations, 
is encoded by a two-block Transformer Encoder (State 
Adapter) into a shared embedding $\hat{e}_s$, which is 
subsequently provided as input to the reward-normalized 
source critic $V^s_\phi(s)$. Lateral tap-point 
connections at layers $L_1, L_2, L_3$ propagate 
weighted representations ($\alpha_k$) from the source 
actors to the target actor $\pi^i_t$ and, analogously, 
propagate another set of lateral representations from 
the source critics, with weights ($\beta_k$), to the 
target critic $V^t_\phi(s)$. The target actor and 
critic are both trained on the original target-domain 
inputs. }
    \label{fig:ASALT_New}
\end{figure}

An overview of the proposed ASALT framework is presented in Figure \ref{fig:ASALT_New}. The central concept is to enable lateral knowledge transfer from frozen source agents to target agents by employing adapter modules that map observations and the global state in the target domain into a shared embedding space, which is subsequently provided as input to the source agents. It contains three important components (i) Observation adapter, (ii) State adapter and (iii) Transfer module

\subsection{Observation adapter} 
As mentioned in introduction, this module maps target-domain observations into a shared embedding space. Learning high-quality embeddings is essential for mitigating negative transfer and facilitating efficient knowledge transfer. The adapter additionally resolves mismatches between source and target observation dimensionalities; even when these dimensionalities are aligned, it can still enhance transfer effectiveness. 

\subsubsection{Fully connected network}

A multi-layer fully connected network maps target-domain 
observations from dimension $o_t$ to $o_s$. For N source 
policies, N independent networks are constructed, each 
generating embeddings aligned to its corresponding source 
policy's input space. This architecture is computationally straightforward; however, because transfer is performed independently for each agent, it may yield suboptimal embeddings by failing to capture inter-agent coordination. To overcome this limitation, we propose to design attention-based adapter.
\subsubsection{Attention Mechanism}

We decompose each target agent’s observation into (i) agent-specific features and (ii) contextual features that describe other agents. For instance, in SMAC the observation comprises both the agent’s own condition and information about allied and enemy units, which serves as contextual input. We denote these components by $\mathbf{o}_a$ and $\mathbf{o}_c = {\mathbf{o}_c}_1, {\mathbf{o}_c}_2, \ldots, {\mathbf{o}_c}_{(M-1)}$, respectively.

The agent-specific component $\mathbf{o}_a$ is processed by a fully connected layer to obtain an embedding $\mathbf{e}_a$. In parallel, the contextual observations $\mathbf{o}_c$ are provided as an input to a Hierarchical Multi-Head Attention (H-MHA) module \citep{paper:hmha}, which applies attention over structured property groups across two 
attention heads $h_1$ and $h_2$ to capture higher-order relational dependencies, producing embeddings $\hat{e}_a$ 
and $e_m$. Here, $\hat{e}_a$ captures agent-level 
relational context, while $e_m$ aggregates the contextual 
features into a compact summary representation. The three 
embeddings are concatenated to form the shared observation 
embedding $\hat{e}_o$, such that:
\begin{equation}
    \hat{e}_o = [e_a \| \hat{e}_a \| e_m]
\end{equation}
The dimensionalities are chosen to be compatible with the 
source policy input space, such that:
\begin{equation}
    \dim(e_a) + \dim(\hat{e}_a) + \dim(e_m) = \dim(\hat{e}_o)
\end{equation}
Similar to the previous setting, we instantiate independent adapters 
for each source policy, where each adapter captures 
target-domain coordination information and maps it 
into the corresponding source policy's embedding space. 

\subsection{State adapter}

Mirroring the observation adapter, the state adapter 
establishes a mapping of the global state between the 
source and target domains when such information is 
available. Concretely, we adopt a two-layer Transformer 
encoder to model the state space of the target domain. 
The target-domain global state $s_t$ is first projected 
into a sequence of token embeddings via a fully connected 
layer, after which these embeddings are processed by the 
Transformer encoder. Each encoder layer comprises a 
Multi-Head Self-Attention (MHSA) sublayer followed by a 
position-wise Feed-Forward Network (FFN), with residual 
connections and layer normalization applied after each 
sublayer. The encoder yields contextualized embeddings 
that form the shared state embedding $\hat{e}_s$, which 
is subsequently provided to the source critics. By 
explicitly capturing interactions among state components 
through self-attention, the proposed design promotes the 
transfer of coordination patterns across domains.

In the same vein, independent state adapters can be 
instantiated for each source critic, enabling each 
adapter to capture target-domain global state information 
and project it into the corresponding source critic's 
embedding space.

\subsection*{Training process}

We investigate two training paradigms for the state and observation adapters: (i) training the adapters independently of the target agent, and (ii) training them jointly with the target agent.

In the independent-training paradigm, we first collect trajectories in both the source and target domains and use these data exclusively to train the adapters. After this adapter-training phase, we freeze the adapter parameters and subsequently train the target agent on top of the learned representations. In the joint-training paradigm, by contrast, the adapters and the target agent are optimized concurrently using trajectories collected in the target domain.

Joint training is typically more sample-efficient than independent training, as it induces target-domain embeddings that explicitly maximize the utility of information transferred from the source agents to the target domain. We provide an empirical comparison of these two training paradigms in Section~\ref{sec:ablation}.

When the source policies operate on observations with heterogeneous dimensionalities, we instantiate a distinct observation adapter for each source policy i.e. we train multiple policy-specific observation adapters rather than a single shared adapter.

\subsection{Transfer module}

In this module, the target-domain agent is trained by leveraging information from a collection of pretrained source agents via lateral knowledge transfer, following the MALT framework \citep{paper:MALT}. Specifically, we introduce layer-wise lateral connections between the target agent and each available source agent. These connections are instantiated for both the actor networks (policy parameterizations) and the critic networks (value estimators), thereby facilitating transfer at the levels of action selection and value-based coordination.

For each target agent, the target actor at a given depth attends to the corresponding representations from all source-domain actors; additionally, attention mechanisms are established between the source and target critics to enable critic-side knowledge transfer. To stabilize critic transfer across domains that may exhibit differing reward scales, rewards are normalized prior to constructing critic-side lateral signals. Information is subsequently propagated laterally through the cross-network connections and integrated within the target network to guide representation learning.

The target agent can be optimized either (i) in an offline manner, using a fixed dataset while maintaining the lateral connections active, or (ii) in an online manner, through direct interaction with the target environment while similarly leveraging the lateral pathways. To mitigate computational complexity, transfer can optionally be constrained to the top $N_s$ source agents (e.g., selected for each target agent according to a predefined similarity or performance criterion), such that only their corresponding lateral contributions are aggregated during training.

\subsection{Training process of ASALT approach}
In this work, we employ Multi-Agent Proximal Policy Optimization (MAPPO) to train the agents. As previously discussed, any method based on the CTDE paradigm can, in principle, be used for agent training. In the MAPPO framework, we consider $N$ actors, one for each agent, parameterized by $\theta_i, \; i = 1, \ldots, N$. In addition, we utilize a shared critic $V_\phi(\mathbf{s})$, parameterized by $\phi$, to estimate the value of the global state $\mathbf{s}$. During the training phase, all $N$ actors and the critic are updated through interactions with the environment. During the execution (deployment) phase, only the $N$ actors are used to compute actions based on their respective local observations. At this stage, we obtain $N$ source agents that can be used for transfer learning.

To train agents in the target domain, we adopt a similar architecture consisting of one actor per agent and a shared critic. During the training, we additionally introduce a weighting module that assigns weights to the outputs of each layer of the source agents and aligns them with the corresponding layers of the target agent. The observation space of the agent in the target domain, $\mathbf{o}_t$, is first processed by an observation adapter $f_o(\cdot)$; the resulting transformed observation is then fed into the source policies, while the original target observation $\mathbf{o}_t$ is provided directly to the target policy. Similarly, the global state of the target domain, $\mathbf{s}_t$, is passed through a state adapter $f_s(\cdot)$, and the transformed state is then used as input to the source critics. To ensure consistent dimensionality and semantic alignment between the source and target domains, we employ the previously introduced observation and state adapters $f_o(\cdot)$ and $f_s(\cdot)$.

It is important to note that no additional action adapter is required to address discrepancies in the dimensionality of action space, in this context, as the final layer of the source policies can be omitted. This layer typically encodes limited information that is relevant for the construction of the target policy and therefore does not substantially contribute to its performance.

\begin{figure}[t]
    \centering
    \subfloat[3m to 8m]{
        \includegraphics[width=0.32\textwidth]{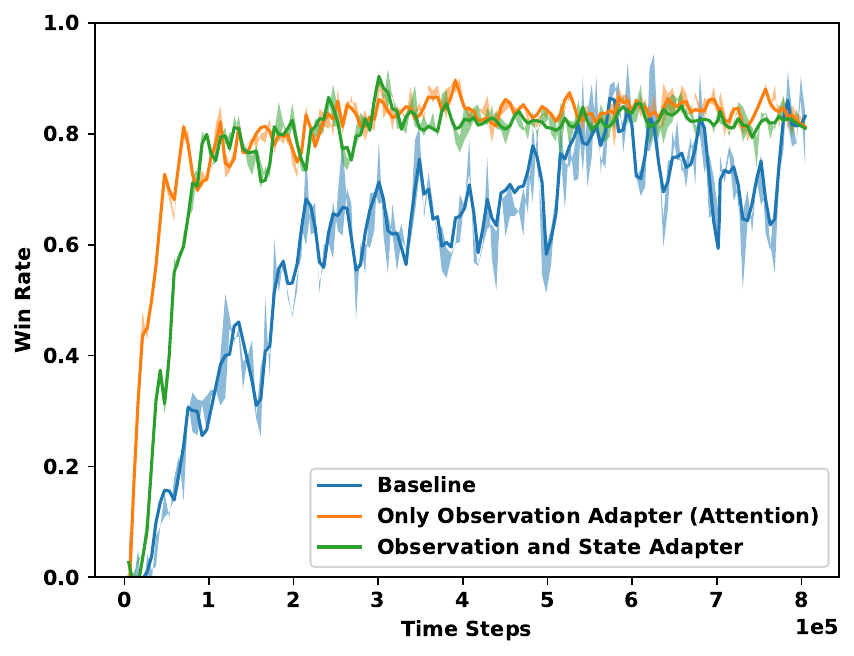}
        \label{fig:smac1}}
    \subfloat[8m to 3m]{
        \includegraphics[width=0.32\textwidth]{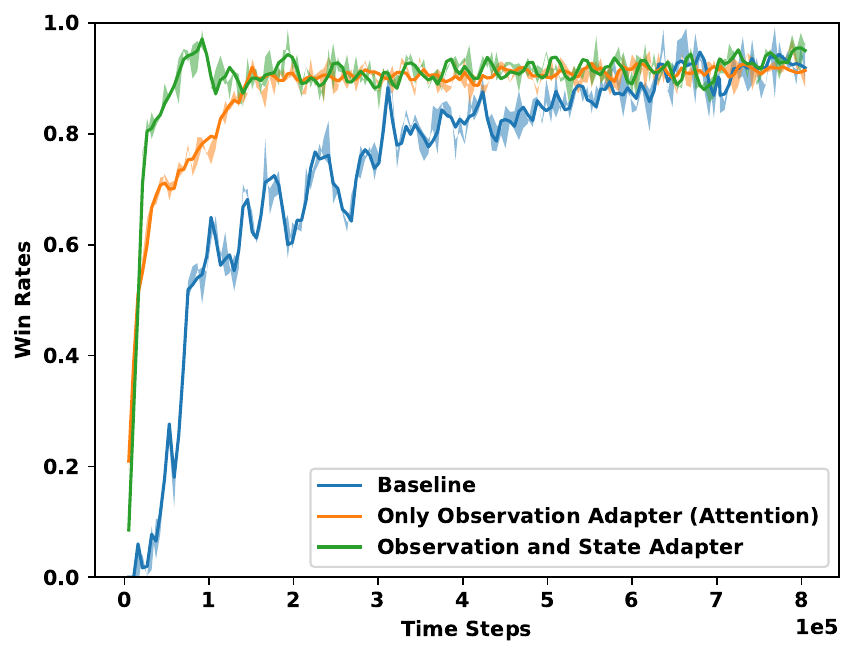}
        \label{fig:smac2}}
    \subfloat[3m to 3s5zvs3s6z]{
        \includegraphics[width=0.32\textwidth]{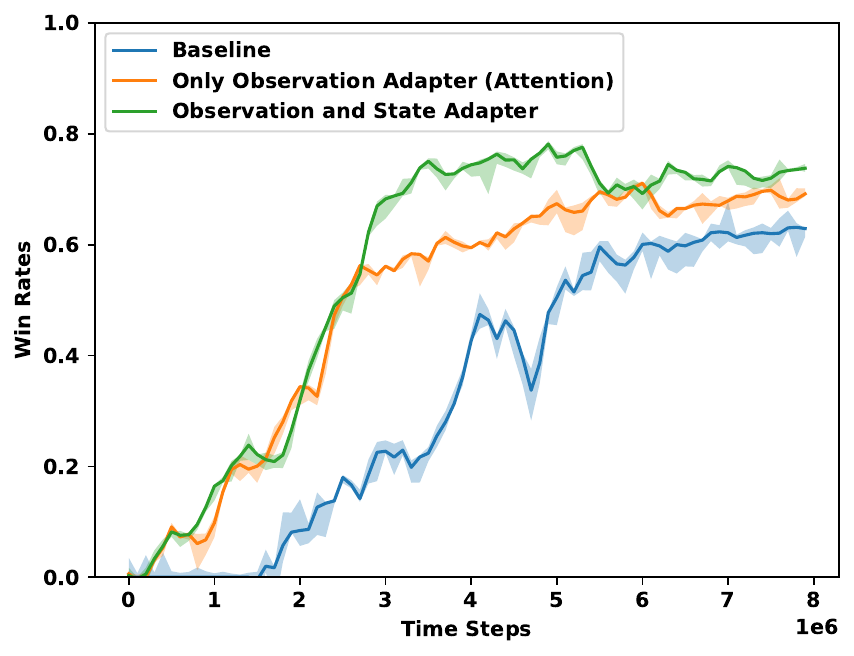}
        \label{fig:smac3}}
    \caption{Win rate as a function of environment interaction 
    steps for transfer learning in SMAC. (a) Up-scaling: 
    $3m \rightarrow 8m$; (b) Down-scaling: $8m \rightarrow 3m$; 
    (c) Heterogeneous transfer: $3m \rightarrow 3s5zvs3s6z$. 
    The ASALT method equipped with both state and observation 
    adapters (green curve) consistently outperforms ASALT using 
    only an observation adapter (orange curve) as well as the 
    baseline method (blue curve). Shaded regions indicate 95\% 
    confidence intervals computed over 10 random seeds.}
    \label{fig:SMAC}
\end{figure}

\section{Results and Discussions}
\label{sec:Results}

To evaluate the efficacy of the proposed ASALT approach, we conducted experiments in three distinct multi-agent environments: (i) the StarCraft II environment (SMAC) \citep{paper:smac}, (ii) the Google Research Football environment, \citep{paper:google}, and (iii) the Multi-Particle Environment (MPE) \citep{paper:MPE}. In SMAC, we considered multiple map configurations spanning a spectrum of difficulty levels from easy to extremely hard, and used these to perform transfer learning.

In the MPE setting, we selected two canonical scenarios: (i) Simple Spread and (ii) Simple Tag. In the Google Football environment, we examined transfer both within full-game scenarios and between academy (training) scenarios and full-game scenarios.

Returning to the discussion, the source agents in all the environments were obtained using MAPPO \citep{paper:MAPPO} across the aforementioned domains. The MAPPO hyperparameters were chosen in a domain-specific manner using conventional tuning practices, based on the characteristics of each environment. We then evaluated ASALT under the following three transfer scenarios:

\begin{enumerate}
    \item Transfer from a source domain to a target domain where the dimensionality of the observation space differs.
    \item Transfer from a source domain to a target domain where the dimensionality of the observation space is identical.
    \item Transfer from a source domain to a target domain in the presence of negative transfer effects.
\end{enumerate}

\begin{table}[t]
\centering
\caption{Comparison of the number of environment interactions 
required to attain 80\% of the asymptotic reward for transfer 
scenarios with observation-space mismatch. ASALT consistently 
outperforms LA-QTransformer and domain adaptation approaches 
(DANN, CORAL, CycleGAN), demonstrating superior sample efficiency 
in multi-agent transfer settings. In addition, ASALT maintains an 
advantage over the LA-QTransformer (SOTA method) across all 
evaluated conditions. The domain adaptation methods exhibit 
comparatively poor performance, likely because they are primarily 
designed for single-agent settings and therefore do not transfer 
effectively to multi-agent domains.}
\label{tab:domain_adaptation}
\begin{adjustbox}{width=\textwidth}
\begin{tabular}{lcccccc}
\hline
\textbf{Scenario} &
\textbf{ASALT} &
\textbf{DANN} &
\textbf{CORAL} &
\textbf{CycleGAN} &
\textbf{LA-QTransformer} &
\textbf{Baseline} \\ \hline
& \multicolumn{6}{c}{\textit{(Environment steps $\times 10^5$)}} \\ \hline
\textbf{3m to 8m} & 
$\mathbf{0.54\pm0.05}$ & 
$3.1\pm0.08$ & 
$2.8\pm0.06$ & 
$4.1\pm0.05$ & 
$1.72\pm0.06$ & 
$5.1\pm0.03$ \\
\textbf{8m to 3m} & 
$\mathbf{0.47\pm0.03}$ & 
$2.8\pm0.06$ & 
$3.3\pm0.07$ & 
$3.1\pm0.04$ & 
$1.42\pm0.02$ & 
$4.82\pm0.03$ \\ \hline
& \multicolumn{6}{c}{\textit{(Environment steps $\times 10^6$)}} \\ \hline
\textbf{3m to 3s5zvs3s6z} & 
$\mathbf{2.85\pm0.02}$ & 
$8.17\pm0.03$ & 
$7.14\pm0.04$ & 
$10.15\pm0.05$ & 
$4.17\pm0.04$ & 
$5.84\pm0.02$ \\
\textbf{3m to 10mvs11m} & 
$\mathbf{0.38\pm0.03}$ & 
$3.67\pm0.05$ & 
$4.23\pm0.03$ & 
$5.10\pm0.04$ & 
$1.15\pm0.04$ & 
$1.25\pm0.01$ \\ \hline
\end{tabular}
\end{adjustbox}
\end{table}

\begin{table}[t]
\centering
\caption{Final win rates at convergence for scenarios with 
mismatched observation spaces. ASALT consistently achieves 
higher performance than both domain adaptation baselines and 
LA-QTransformer, indicating more effective transfer of 
coordination knowledge.}
\label{tab:domain_adaptation_1}
\begin{adjustbox}{width=\textwidth}
\begin{tabular}{lcccccc}
\hline
\textbf{Scenario} &
\textbf{ASALT} &
\textbf{DANN} &
\textbf{CORAL} &
\textbf{CycleGAN} &
\textbf{LA-QTransformer} &
\textbf{Baseline} \\ \hline
\textbf{3m to 8m} & 
$\mathbf{0.82\pm0.04}$ & 
$0.79\pm0.06$ & 
$0.81\pm0.04$ & 
$0.75\pm0.05$ & 
$0.81\pm0.03$ & 
$0.8\pm0.02$ \\
\textbf{8m to 3m} & 
$0.94\pm0.03$ & 
$0.92\pm0.04$ & 
$0.94\pm0.01$ & 
$0.91\pm0.03$ & 
$0.94\pm0.02$ & 
$\mathbf{0.96\pm0.02}$ \\
\textbf{3m to 3s5zvs3s6z} & 
$\mathbf{0.72\pm0.04}$ & 
$0.61\pm0.06$ & 
$0.54\pm0.04$ & 
$0.65\pm0.03$ & 
$0.62\pm0.04$ & 
$0.66\pm0.03$ \\
\textbf{3m to 10mvs11m} & 
$\mathbf{0.8\pm0.03}$ & 
$0.65\pm0.05$ & 
$0.68\pm0.08$ & 
$0.66\pm0.03$ & 
$0.75\pm0.05$ & 
$0.77\pm0.06$ \\ \hline
\end{tabular}
\end{adjustbox}
\end{table}

\subsection{Baselines}
We compare performance across two categories of methods: (i) domain adaptation approaches—DANN \citep{paper:DANN}, CORAL \citep{paper:CORAL}, and CycleGAN \citep{paper:CycleGAN}; and (ii) transfer learning approaches—MALT \citep{paper:MALT}, PSMARL \citep{paper:PSMARL}, Policy Distillation \citep{paper:Distill}, EPC \citep{paper:EPC}, LA-QTransformer \citep{paper:MARL_Trasnformer_Baseline}, and a fine-tuned policy.

The first category of methods is capable of accommodating mismatches in the dimensionality of the observation space. However, these approaches were originally developed for single-agent reinforcement learning. We therefore extended them to the multi-agent setting using a straightforward agent-wise transfer scheme. Specifically, we employed a clustering-based procedure to identify, for each target-domain agent, a corresponding similar agent in the source domain to which transfer can be applied.

The second category of methods natively supports multi-agent scenarios but is unable to handle changes in the dimensionality of the observation space between source and target domains. Although LA-QTransformer \citep{paper:MARL_Trasnformer_Baseline} can accommodate mismatches in observation-space dimensionality, it does not explicitly address the issue of negative transfer. Finally, we report results relative to a baseline in which agents are trained from scratch using MAPPO.

\subsection{Metrics for comparison}
We compare our results with those of the baseline methods using two evaluation metrics: (i) the time required to reach \(80\%\) of the final converged reward, and (ii) the final value of the converged reward. The first metric provides an indication of the learning speed of the system, while the second reflects its ultimate performance. All values reported in the tables are expressed as mean \(\pm\) standard deviation over 10 random seeds. Convergence time in the results is measured in environment steps. The final reward is computed as the average win rate over the last 100 evaluation episodes.

In the following, we report a comparative analysis of the results obtained across different map scenarios in SMAC against the aforementioned baseline methods.

\subsection{Scenario 1} The plots corresponding to this scenario are presented in Figures \ref{fig:SMAC}. In Figure \ref{fig:smac1}, the source domain is the $3m$ map in the SMAC environment, which comprises $3$ allied units and an equal number of enemy units, whereas the target domain is the $8m$ map, consisting of $8$ allies and $8$ enemies. These results indicate that the proposed ASALT methods consistently outperform the baseline. The same figure also reports results for a downscaling task, in which the target domain contains fewer agents than the source domain, as well as for a heterogeneous scenario, in which the agent types differ between source and target domains. In these settings as well, ASALT exhibits superior performance relative to training from scratch. Moreover, it is evident that ASALT equipped with both state and observation adapters attains better performance than configurations relying solely on observation adapters. A more fine-grained comparison is provided in the ablation study section.

We additionally compare ASALT against domain adaptation methods, which can in principle address state mismatches but are not designed for multi-agent systems. To apply these methods in the multi-agent setting, we pair each target-domain agent with the most similar source-domain agent, where similarity is assessed based on experience collected under a random policy, and then perform agent-to-agent transfer. The corresponding results are reported in Tables \ref{tab:domain_adaptation} and \ref{tab:domain_adaptation_1}. These values show that the proposed ASALT framework performs favorably relative to existing baselines. This performance gain can be attributed to the fact that the state adapter transfers coordination knowledge rather than attempting a direct state-to-state mapping, thereby enabling a more effective transfer of learned behaviors. In particular, the ASALT variant that transfers both policy and critic knowledge outperforms the ASALT variant that transfers from the policy alone. This improvement arises because the joint transfer of policies and critics facilitates more accurate estimation of the advantage function using knowledge acquired from the source policies.

\subsection{Scenario 2} In this experimental setting, we evaluate the ASALT approach against established techniques, specifically transfer-learning baselines in multi-agent systems. Our objective is to demonstrate the effectiveness of ASALT relative to these baselines in scenarios where the source and target domains share an identical observation space dimensionality.

\begin{table}[t]
\centering
\caption{Environment steps required to attain 80\% of the final 
reward are reported for scenarios in which the dimensionality of 
the observation space remains constant. ASALT exhibits more rapid 
convergence relative to multi-agent transfer baselines (MALT, 
PSMARL, Policy Distillation, EPC, and fine-tuned policies). This 
superior performance can be attributed to the ASALT-transformer's 
mechanism of transferring coordination strategies to the target 
agents. By focusing on coordination transfer, ASALT 
achieves performance that surpasses existing approaches.}
\label{tab:same_observation}
\begin{adjustbox}{width=\textwidth}
\begin{tabular}{lccccccc}
\hline
\textbf{Scenario} &
\textbf{ASALT} &
\textbf{MALT} &
\textbf{PSMARL} &
\textbf{Distilled Policy} &
\textbf{EPC} &
\textbf{Fine-tune} &
\textbf{Baseline} \\ \hline
& \multicolumn{7}{c}{\textit{(Environment steps $\times 10^5$)}} \\ \hline
\textbf{3svs3z to 3svs4z} & 
$\mathbf{0.62\pm0.02}$ & 
$1.17\pm0.04$ & 
$3.2\pm0.04$ & 
$2.59\pm0.04$ & 
$1.29\pm0.05$ & 
$2.91\pm0.04$ & 
$3.72\pm0.02$ \\
\textbf{3svs3z to 3svs5z} & 
$\mathbf{0.89\pm0.02}$ & 
$1.41\pm0.03$ & 
$4.05\pm0.05$ & 
$3.42\pm0.04$ & 
$2.18\pm0.04$ & 
$3.92\pm0.03$ & 
$4.17\pm0.03$ \\ \hline
& \multicolumn{7}{c}{\textit{(Environment steps $\times 10^6$)}} \\ \hline
\textbf{8mvs8m to 8mvs9m} & 
$\mathbf{0.99\pm0.04}$ & 
$2.41\pm0.05$ & 
$5.23\pm0.07$ & 
$4.28\pm0.03$ & 
$3.02\pm0.04$ & 
$5.43\pm0.05$ & 
$5.84\pm0.02$ \\ \hline
\end{tabular}
\end{adjustbox}
\end{table}

The results obtained for this scenario are presented in Tables \ref{tab:same_observation} and \ref{tab:same_observation_1}. These results indicate that the proposed ASALT framework consistently outperforms existing techniques, including MALT and curriculum learning. The superior performance of ASALT relative to MALT can be attributed to the use of coordination transfer rather than direct state-to-state transfer, thereby enabling a more effective and efficient knowledge transfer from the source to the target domain. 

\begin{table}[t]
\centering
\caption{Final win rates for scenarios without observation-space 
mismatch indicate that ASALT attains performance comparable to 
state-of-the-art multi-agent transfer baselines, thereby 
empirically demonstrating its robustness in coordination transfer.}
\label{tab:same_observation_1}
\begin{adjustbox}{width=\textwidth}
\begin{tabular}{lccccccc}
\hline
\textbf{Scenario} &
\textbf{ASALT} &
\textbf{MALT} &
\textbf{PSMARL} &
\textbf{Distilled Policy} &
\textbf{EPC} &
\textbf{Fine-tune} &
\textbf{Baseline} \\ \hline
\textbf{3svs3z to 3svs4z} & 
$\mathbf{0.95\pm0.01}$ & 
$0.94\pm0.03$ & 
$0.91\pm0.06$ & 
$0.86\pm0.05$ & 
$0.91\pm0.04$ & 
$0.92\pm0.02$ & 
$0.95\pm0.03$ \\
\textbf{3svs3z to 3svs5z} & 
$\mathbf{0.96\pm0.02}$ & 
$0.89\pm0.05$ & 
$0.88\pm0.06$ & 
$0.92\pm0.03$ & 
$0.94\pm0.02$ & 
$0.91\pm0.03$ & 
$\mathbf{0.96\pm0.02}$ \\
\textbf{8mvs8m to 8mvs9m} & 
$\mathbf{0.96\pm0.03}$ & 
$0.89\pm0.04$ & 
$0.91\pm0.05$ & 
$0.86\pm0.05$ & 
$0.89\pm0.03$ & 
$0.91\pm0.04$ & 
$0.94\pm0.03$ \\ \hline
\end{tabular}
\end{adjustbox}
\end{table}

It should be noted that, although curriculum transfer achieves superior performance compared to MALT, it is computationally much more expensive. Consequently, we have selected MALT as the baseline method for evaluating the ASALT approach. Nonetheless, it is important to emphasize that the proposed state-adapters are model-agnostic and can be integrated into any underlying method.

\begin{table}[t]
\centering
\caption{Environment interaction steps required to attain 80\% of 
the asymptotic return in negative transfer scenarios. The ASALT 
approaches attenuate negative transfer effects relative to both 
domain adaptation baselines and the standard baseline method. By 
constructing and leveraging state embeddings instead of raw 
observations, and by transferring these embeddings to the target critics, ASALT selectively transfers only the task-relevant 
information rather than the full observation space. This targeted 
transfer mechanism results in a measurable reduction of negative 
transfer.}
\label{tab:negative_transfer}
\begin{adjustbox}{width=\textwidth}
\begin{tabular}{lcccccc}
\hline
\textbf{Scenario} &
\textbf{ASALT} &
\textbf{DANN} &
\textbf{CORAL} &
\textbf{CycleGAN} &
\textbf{LA-QTransformer} &
\textbf{Baseline} \\ \hline
& \multicolumn{6}{c}{\textit{(Environment steps $\times 10^5$)}} \\ \hline
\textbf{3m to 3s5zvs3s6z} & 
$\mathbf{2.9\pm0.03}$ & 
$6.17\pm0.06$ & 
$5.2\pm0.07$ & 
$5.59\pm0.04$ & 
$6.92\pm0.05$ & 
$5.84\pm0.02$ \\ \hline
& \multicolumn{6}{c}{\textit{(Environment steps $\times 10^6$)}} \\ \hline
\textbf{\begin{tabular}[l]{@{}l@{}}10m\_vs\_11m\\ to\\ corridor\end{tabular}} & 
$\mathbf{2.89\pm0.03}$ & 
$7.87\pm0.04$ & 
$9.24\pm0.05$ & 
$9.21\pm0.07$ & 
$11.81\pm0.06$ & 
$8.87\pm0.01$ \\ \hline
\end{tabular}
\end{adjustbox}
\end{table}

\begin{table}[t]
\centering
\caption{Final win rates in negative transfer scenarios. ASALT 
surpasses domain adaptation baselines, highlighting the 
effectiveness of embedding-based coordination transfer in 
alleviating negative transfer effects.}
\label{tab:negative_transfer_1}
\begin{adjustbox}{width=\textwidth}
\begin{tabular}{lcccccc}
\hline
\textbf{Scenario} &
\textbf{ASALT} &
\textbf{DANN} &
\textbf{CORAL} &
\textbf{CycleGAN} &
\textbf{LA-QTransformer} &
\textbf{Baseline} \\ \hline
\textbf{3m to 3s5zvs3s6z} & 
$\mathbf{0.69\pm0.03}$ & 
$0.61\pm0.05$ & 
$0.62\pm0.07$ & 
$0.59\pm0.08$ & 
$0.46\pm0.04$ & 
$0.68\pm0.07$ \\
\textbf{\begin{tabular}[l]{@{}l@{}}10mvs11m \\ to \\ corridor\end{tabular}} & 
$\mathbf{0.89\pm0.04}$ & 
$0.77\pm0.05$ & 
$0.75\pm0.06$ & 
$0.76\pm0.07$ & 
$0.71\pm0.05$ & 
$\mathbf{0.89\pm0.04}$ \\ \hline
\end{tabular}
\end{adjustbox}
\end{table}

\subsection{Scenario 3} In this set of experiments, we evaluate the proposed approach with respect to its ability to mitigate negative transfer. To this end, we consider multiple map configurations in SMAC that demand qualitatively different forms of coordination. For instance, in the $3m$ map, the optimal policy is characterized by a hit-and-run maneuver, whereas in the $3s$vs$3z$ map, agents must employ a distinct coordination pattern involving focused fire and context-dependent retreat. Consequently, these two maps exemplify coordination requirements that are not only different but potentially conflicting. Analogously, the broader SMAC environment encompasses several map scenarios, each inducing its own coordination regime.

We design two sets of experiments specifically to study negative transfer phenomena arising between such heterogeneous map scenarios. The corresponding performance results, along with those of the comparative methods, are reported in Tables \ref{tab:negative_transfer} and \ref{tab:negative_transfer_1}. As shown, the proposed ASALT method consistently outperforms both existing domain-adaptation-based baselines and the non-adaptive baseline. This superior robustness to negative transfer constitutes an additional advantage of ASALT, enhancing its suitability for deployment in real-time applications.

We observe that in the transfer from 3m $\rightarrow$ 8m transfer setting, actor-only transfer slightly outperforms actor–critic transfer in final performance. This can be attributed to the substantial increase in state-space dimensionality and the resulting need for re-learning coordination strategies. In such cases, the value estimates transferred via the critic may have limited applicability, whereas policy representations can adapt more flexibly. Nevertheless, actor–critic transfer still provides improved stability and faster initial learning.
\subsection{Football Environment}

As previously discussed, we evaluated the proposed ASALT approach in the Google Research Football (GRF) environment. This environment comprises two classes of scenarios: (i) academy scenarios and (ii) full-game scenarios. We investigated transfer both within a full-game scenario and between academy and full-game scenarios.

The corresponding results are reported in Table \ref{tab:Google}. In the first setting, we examined transfer from a $5$-versus-$5$ to an $11$-versus-$11$ scenario. This represents an upscaling problem in which the number of agents increases from $5$ to $11$, and the number of opponents likewise increases from $5$ to $11$, thereby inducing qualitatively different coordination requirements. The results indicate that the proposed ASALT variant employing a transformer backbone requires less than $75\%$ of the time taken by the baseline to achieve $80\%$ of the final converged reward. 

\begin{table}[t]
\centering
\caption{Performance of ASALT methods in the Google Research 
Football environment under varying experimental conditions. 
Across scenarios characterized by observation space mismatch, 
invariant state dimensionality, and the presence of negative 
transfer, ASALT demonstrates consistently faster convergence 
relative to the baseline approaches.}
\label{tab:Google}
\begin{tabular}{lcc}
\hline
\textbf{Scenario} &
\textbf{ASALT} &
\textbf{Baseline} \\ \hline
& \multicolumn{2}{c}{\textit{(Environment steps $\times 10^8$)}} \\ \hline
\textbf{5vs5 to 11vs11} & 
$\mathbf{0.38\pm0.0004}$ & 
$0.77\pm0.001$ \\
\textbf{\begin{tabular}[l]{@{}l@{}}3vs1 with keeper \\ to 5vs5 (Full game)\end{tabular}} & 
$\mathbf{0.11\pm0.0005}$ & 
$0.4\pm0.003$ \\ \hline
\end{tabular}
\end{table}

We additionally evaluated a negative transfer setting, in which knowledge is transferred from an academy scenario (3-versus-1 with goalkeeper) to a full-game scenario (5-versus-5). In this case, the agents must adapt and modify the coordination patterns acquired in the source task. Here as well, the ASALT approach converges more rapidly than the baseline, achieving a reduction in training time of approximately $70\%$. Overall, these results demonstrate that the proposed ASALT framework attains faster convergence relative to the baseline, even under negative transfer conditions.

\subsection{MPE}
The results obtained in the MPE environment for the different scenarios are presented in Table \ref{tab:MPE}. The first row of the table corresponds to the setting in which there is a mismatch in the observation space between the source and target domains. The second row corresponds to the setting in which no such mismatch is present. The final row represents the scenario requiring a different form of coordination between the source and target domains (i.e., a negative transfer setting).

From the values reported in the table, the proposed ASALT approach consistently outperforms the baseline methods, indicating faster convergence relative to the baselines. The final converged reward is not reported, as the asymptotic performance is comparable across all methods.

\begin{table}[t]
\centering
\caption{Performance of ASALT methodologies on the MPE benchmark 
environments. ASALT exhibits accelerated convergence relative to 
baseline methods across scenarios characterized by observation 
space mismatch, invariant dimensionality, and negative transfer 
phenomena.}
\label{tab:MPE}
\begin{tabular}{lcc}
\hline
\textbf{Scenario} &
\textbf{ASALT} &
\textbf{Baseline} \\ \hline
& \multicolumn{2}{c}{\textit{(Environment steps)}} \\ \hline
\textbf{\begin{tabular}[l]{@{}l@{}}Simple Spread from \\ 2 to 5 Landmarks\end{tabular}} &
$\mathbf{324\pm21}$ &
$2145\pm41$ \\
\textbf{\begin{tabular}[l]{@{}l@{}}Simple Tag with prey \\ speed increased from 1 to 2\end{tabular}} &
$\mathbf{4749\pm51}$ &
$22247\pm132$ \\
\textbf{Simple Spread to Simple Tag} &
$\mathbf{8487\pm69}$ &
$18147\pm123$ \\ \hline
\end{tabular}
\end{table}

\subsection{Ablation Studies}
\label{sec:ablation}

We have tested three ablation studies in this work (i) Training mechanism: Joint vs independent training of adapters, (ii) With and without state adapters (with different types of observation adapters) and (iii) Number of source agents to transfer (Given as supplementary material). 

\subsubsection{Training mechanism}
For the ablation study, the observation and state adapter modules were trained independently of the target agent. Specifically, we first collected sample trajectories in both the source and target domains and subsequently trained the state and observation adapters using these datasets. After training, the parameters of the adapters were held fixed while optimizing the target agent. The experiments were conducted on cross-scenario transfer between the $3m$ and $8m$ map configurations in SMAC, in both directions.

The outcomes of this study are presented in Table \ref{tab:ablation}. The results demonstrate that joint training of the adapters and the agent leads to markedly improved coordination performance compared to the independent training paradigm.

\begin{table}[t]
\centering
\caption{Ablation analysis comparing joint training and independent 
training of the state adapter module demonstrates that joint 
optimization substantially accelerates convergence relative to 
both independently trained variants and the baseline model. These 
findings empirically substantiate the critical role of integrated 
optimization in enhancing training efficiency and overall model 
performance.}
\label{tab:ablation}
\begin{tabular}{lccc}
\hline
\textbf{Scenario} & 
\textbf{Joint Training} & 
\textbf{Independent Training} & 
\textbf{Baseline} \\ \hline
& \multicolumn{3}{c}{\textit{(Environment steps $\times 10^5$)}} \\ \hline
\textbf{3m to 8m} & 
$\mathbf{0.72\pm0.04}$ & 
$3.16\pm0.06$ & 
$5.1\pm0.03$ \\
\textbf{8m to 3m} & 
$\mathbf{0.57\pm0.02}$ & 
$2.54\pm0.04$ & 
$4.82\pm0.03$ \\ \hline
\end{tabular}
\end{table}

\subsubsection{Transfer of both actor and critic vs only actor}
In this ablation study, we evaluate the performance of the proposed ASALT approach with and without state adapters i.e. transfer happens only with actors vs both actor and critic. The corresponding comparison is presented in Figure \ref{fig:SMAC_new}. From this figure, it can be observed that the performance improves relative to the configuration that employs only observation adapters. This improvement can be attributed to enhanced agent learning, as the additional state information enables the critics to compute more accurate advantage estimates.

Furthermore, we compare two variants of observation adapters: (i) ANN and (ii) an attention-based network. The attention-based observation adapters exhibit superior performance compared to the ANN-based methods. This performance gain is particularly pronounced in scenarios involving upscaling (i.e., transferring knowledge from a smaller number of agents to a larger number of agents). In such settings, attention mechanisms facilitate more effective transfer of coordination patterns among agents, leading to improved overall performance.

\subsection{Computational Time}  Because the additional state/observation adapter component(s) are trained jointly with the target agent, the proposed approach incurs only marginal computational overhead relative to existing methods. For the $3m \rightarrow 8m$ map scenario, using $3$ source policies and $1$ source critic, together with $3$ observation adapters and $1$ state adapter, the proposed ASALT method required $27$ hours to converge, whereas the corresponding configuration without any state adapters required $24$ hours. This increase in computational time is relatively minor when contrasted with the accelerated convergence achieved by the proposed approach. In the absence of transfer learning, training the target agent from scratch required $49$ hours to converge. On average, the ASALT framework resulted in a $21\%$ increase in computational time. All experiments were conducted on an Apple M1 GPU-equipped system with 64 GB of memory. Furthermore, in line with other lateral transfer techniques, the proposed method scales to arbitrarily large multi-agent systems.

\section{Conclusions}

In this work, we have proposed a way to perform efficient transfer in MARL where there is a mismatch in the observation space between source and target domain. ASALT, which uses a state adapters to transfer the coordination instead of transferring them to observation spaces. This design resulted in more efficient transfer and reduces the impact of the negative transfer. Experiments on SMAC and MPE environments shows that the ASALT has shown improved performance over existing benchmarks. While critic-side transfer contributes to improved stability and faster early learning in several cases, it is not universally optimal. In scenarios involving large changes in team size or state-space dimensionality, actor-only transfer can sometimes achieve comparable or better asymptotic performance.

Future studies involve theoretical analysis of the proposed approach along with designing hierarchal transfer between domains for further reduce the computational cost.


\appendix

\bibliography{main}
\bibliographystyle{rlj}

\end{document}